%% file: TSA-Main.tex
\documentclass[sigconf]{acmart}
\usepackage{graphicx}
\usepackage{booktabs} % For formal tables
\usepackage[tight]{subfigure}
\usepackage{xspace}
\usepackage{multirow} % for tables
\usepackage{array} % for tables, required for tabularx
\usepackage{tabularx} % for tables
\usepackage{enumitem}

\graphicspath{{Figures/}}

\newcommand{\ie}{\emph{i.e.,}\xspace}
\newcommand{\eg}{\emph{e.g.,}\xspace}

\newcommand{\paratitle}[1]{\vspace{1ex}\noindent \textbf{#1}}

\setcopyright{rightsretained}
\acmDOI{}

\acmISBN{}

\acmConference[Draft]{Sentiment Analysis}{2019}{Sentiment Analysis}
\begin{document}

\title{Targeted Sentiment Analysis: A Data-Driven Categorization}
% Predicting Stability of Clinical Concept Embeddings
% Stability of Word2vec for Clinical Concepts: Analysis and Prediction

 \author{Jiaxin Pei}
 \affiliation{%
   \institution{Wuhan University}
   \city{Hubei, China}
  }
 \email{pedropei@whu.edu.cn}

 \author{Aixin Sun}
 \affiliation{%
   \institution{Nanyang Technological University}
   \city{Nanyang Avenue, Singapore}
  }
 \email{axsun@ntu.edu.sg}

 \author{Chenliang Li}
 \affiliation{%
   \institution{Wuhan University}
   \city{Hubei, China}
 }
 \email{cllee@whu.edu.cn}

\begin{abstract}
Targeted sentiment analysis (TSA), also known as aspect based sentiment analysis (ABSA), aims at detecting fine-grained sentiment polarity towards targets in a given opinion document.  Due to the lack of labeled datasets and effective technology, TSA had been intractable for many years. The newly released datasets and the rapid development of deep learning technologies are key enablers for the recent significant progress made in this area. However, the TSA tasks have been defined in various ways with different understandings towards basic concepts like ``target'' and ``aspect''. In this paper, we categorize the different tasks and highlight the differences in the available datasets and their specific tasks. We then further discuss the challenges related to data collection and data annotation which are overlooked in many previous studies.
\end{abstract}



\maketitle

\input{mainContent.tex}

\bibliographystyle{ACM-Reference-Format}
\bibliography{sentiment}

\end{document}

%% file: mainContent.tex
%=====================================
\section{Introduction}
\label{sec:intro}
%=====================================

Sentiment analysis or opinion mining is the computational study of people’s opinions, sentiments, emotions, appraisals, and attitudes towards entities such as products, services, organizations, individuals, issues, events, topics, and their attributes~\cite{Liu2015SentimentAM}. Sentiment analysis could be broadly classified into three categories: \textit{document-level sentiment analysis}, \textit{sentence-level sentiment analysis} and \textit{aspect-level sentiment analysis}~\cite{zhang2018deep}. Document-level and sentence-level sentiment analysis aim to assign sentiment polarity towards a given opinion document at different level of granularity, \ie document level and sentence level respectively. However, neither document-level nor sentence-level sentiment analysis could tell what people like or dislike exactly because both tasks do not aim at recognizing opinion target. For example, given a sentence ``\textit{I like the food here, but the service is terrible.}'',  document-level or sentence-level sentiment analysis could not identify the different attitudes towards``food'' and ``service''.  Aspect-level sentiment analysis can be applied here to identify the aspects and sentiments towards the identified aspects. On the other hand, the above example sentence does not explicitly provide the ``entity'' (\eg a restaurant or a hotel), to which the sentiments are targeted on. In recent studies, researchers have considered both \textit{aspect} and \textit{entity} as the targets of sentiment analysis, and defined the task of \textit{\textbf{targeted sentiment analysis}} (TSA)~\cite{Welch2016TargetedST,Ma2018TargetedAS}.  In simple words, TSA aims to identify sentiment towards each opinion target. However, depending on the meaning of ``target'' (\eg aspect, entity, or entity-aspect pairs), targeted sentiment analysis could refer to various different tasks and has been applied to datasets of different characteristics. In this paper, we aim to provide a categorization of the various tasks under the big umbrella of ``targeted sentiment analysis''. Our categorization of the tasks is mainly based on the characteristics of the datasets used in these different targeted sentiment analysis.

There are two key enablers for the tasks of targeted sentiment analysis as a whole. First is the availability of datasets.  For document- and sentence-level sentiment analysis, product reviews are often the data of study. The star ratings that usually come with product reviews, are considered as indicators of sentiment polarities of the corresponding reviews. However, TSA requires more fine-grained annotations, and such annotations are expensive to collect. In particular, the two datasets released by~\citeauthor{Pontiki2014SemEval} in SemEval 2014 task 4 gained significant interests from the research community. The second key enabler is deep learning technology. Traditional sentiment analysis solutions, including lexicon based approaches and machine learning based approaches, rely heavily on human crafted features~\cite{Jiang2011TargetdependentTS,Lazaridou2013ABM}. Such solutions achieve promising results on document- and sentence-level tasks, but TSA remains challenging. Recently deep learning was introduced to address many TSA tasks~\cite{wang-EtAl:2016:EMNLP20163,zhang2016gated,tang-qin-liu:2016:EMNLP2016}. Neural models could produce better representation of data through multiple layers of non-linear transformation, thus modeling fine-grained interactions between target and context became tangible. More importantly, end-to-end models release people from the boring and time-consuming feature engineering process.

While many deep learning approaches have been proposed to address TSA tasks, we have witnessed confusion of basic concepts like ``target'', ``aspect'' and ``entity'' in recent studies. That is, the definition of TSA tasks varies from one study to another.  The differences in TSA tasks leads to confusion on choice of baseline models for performance evaluation as well as the applicability of models on specific datasets.  Although the definition proposed by~\cite{Liu2015SentimentAM} is broad enough to cover all the new emerging tasks in TSA, the definition is too abstract and does not indicate the detailed and essential differences between variants of TSA tasks reported in recent studies. In this paper, we provide a data-driven categorization of TSA tasks and we briefly review the recent solutions to each of their sub-tasks. We also hope that this categorization could serve as a practical guide for researchers or practitioners  who want to apply TSA in real-world scenarios. Apart from categorization, we also emphasize the issues related to data. Specifically, existing works mainly focus on devising new models based on certain benchmark datasets. However, how to collect appropriate data and the effort of data preparation are often overlooked. The main contributions of this work are two-fold. We survey existing tasks, datasets, and recent solutions for TSA and categorize them into three subcategories.  We propose the data preparation problem from information retrieval perspective which was not well covered in existing literatures.

%=====================================
\section{Definitions and Datasets}
\label{sec:definition}
%=====================================

Many names have been used to refer to targeted sentiment analysis, including Aspect-level Sentiment Analysis~\cite{Steinberger2014AspectLevelSA}, Aspect Based Sentiment Analysis~\cite{wang-EtAl:2016:EMNLP20163}, Feature Based Sentiment Analysis~\cite{erdmann2014feature}, and Topic Based Sentiment Analysis~\cite{Thelwall2013TopicbasedSA}. We use \textit{Targeted Sentiment Analysis} because ``target'' is broad enough to cover relevant concepts including ``entity'' and ``aspect'' and ``entity-aspect'' pairs, to be defined shortly. We mainly follow the definitions of \textit{entity} and \textit{aspect} defined in the  book~\cite{Liu2015SentimentAM}.

\begin{definition}[\textbf{Entity}]
An entity $e$ can be a product, service, topic, person, organization, issue, or an event.
\end{definition}
Note that each entity may have many parts and attributes. In~\cite{Liu2015SentimentAM}, a hierarchical structure ($T$, $W$) is used to describe an entity $e$. $T$ is a hierarchy of parts, subparts, and so on, and $W$ is a set of attributes. However, the hierarchical structures existed in sentences are usually shallow. Further, labeling and detecting complex structures remains extremely challenging. Here, we do not consider complex structures.

\begin{definition}[\textbf{Aspect}]
An aspect $a$ is a part or an attribute of a target entity.
\end{definition}
Aspects may or may not be explicitly mentioned in an opinion document or sentence. In this sentence ``\textit{The food is great but expensive}'', sentiments are expressed on two aspects ``food'' which is explicitly mentioned, and ``price'' which is implicit.

\begin{definition}[\textbf{Target}]
The sentiment target of an opinion is an entity, or a part or attribute of the entity that the sentiment has been expressed upon.
\end{definition}
In targeted sentiment analysis, a sentiment target $t$ could be described by a shallow hierarchical structure ($e$, $a$) where $e$ is the target entity and $a$ refers to an aspect of $e$. Depending on input data,  target entity $e$ and aspect $a$ may or may not be known.
%We will elaborate this point further shortly, based on the datasets used in recent studies.

An \textbf{opinion} in sentiment analysis is a target-sentiment pair ($t, s$), where $t$ is the sentiment target, $s$ is the sentiment expressed on the target $t$ (\eg positive, negative, and neutral). Note that, in the original definition~\cite{Liu2015SentimentAM}, an opinion is described as a quadruple ($t, s, ti, h$) where $h$ is the opinion holder, and $ti$ is the opinion posting time. However, the extraction or identification of opinion holder and posting time is typically not the key focus of TSA, thus we use ($t, s$) to describe opinion for simplicity. Accordingly, an \textit{opinion document} is a document that contains opinions about a finite set of targets $(t_1, t_2,  \ldots, t_n)$.

%===========================================
\subsection{Datasets Used for TSA}
\label{ssec:dataset}
%===========================================
A few datasets have been developed for TSA, particularly aspect-based sentiment analysis. Here, we briefly review five of them which have been used in a number of studies.

\paratitle{SemEval2014 Dataset.}  SemEval 2014~\cite{Pontiki2014SemEval} is the most widely used dataset for aspect-based sentiment analysis.\footnote{SemEval also held TSA tasks in 2015 and 2016. Here we only introduce the most representative task in 2014.} This dataset  contains online reviews for restaurants and laptops, in which each aspect term and its polarity are manually annotated. Each review may consist of multiple sentences and each sentence may contain multiple annotated aspect terms. The aspect terms are from predefined aspect categories such as \textit{``food''}, \textit{``service''}, \textit{``price''},\textit{``ambience''},\textit{``anecdotes/ miscellaneous''}. Their polarities are  assigned to each review.
The possible values of the polarity field are: \textit{``positive''}, \textit{``negative''}, \textit{``neutral''}, and \textit{``conflict''}. However, since very few aspect terms are labeled as \textit{``conflict''}, most of the previous works do not use this label.

\paratitle{Twitter Datasets}. The Twitter dataset collected by~\cite{dong2014adaptive} is an open domain dataset collected through Twitter API using keywords. The \textit{keywords are names} of celebrities, companies, and products. Each tweet is labeled with its polarity towards the keyword entity appearing in the tweet. For example the tweet \textit{``i love britney spears. yes, I said it''} is labeled as positive if a keyword entity ``britney spears'' is used for data collection. Note that, the minimum labeling unit is a tweet and no aspects are labeled. Because a set of targeted keywords is used, if a tweet contains two names but only one of them matches the keyword used for collection, then only the matched name will be annotated.

There is another Twitter dataset~\cite{mitchell2013open} of similar format. This collection contains 7105 Spanish and 2350 English tweets with their entities and targeted sentiment annotated. This dataset was built on tweets collected by~\citeauthor{Etter2012NeritN} for the purpose of studying named entity recognition.  Entities in this dataset are either \textit{Person} or \textit{Organization}. Then the sentiments are further annotated for these entities, and in the labeling process, tweets without consensus on named entities are removed.

\paratitle{BabyCare and Sentihood Datasets.}
Baby Care dataset~\cite{yang2018multi} is probably the most fine-grained dataset used in aspect-based sentiment analysis. This dataset is collected from one of the biggest baby care forums in China.\footnote{\url{http://www.babytree.com}} The main topics discussed in this forums include diaper, milk powder, baby health, and other domain-specific topics. In a given sentence, more than one (entity-aspect:polarity) pair could be annotated where the entities are products (\eg Kao diaper) and aspects are predefined (\eg anti-leakage, anti-allergy, price). An example annotation is  (Kao diaper - price): negative.

Sentihood~\cite{Saeidi2016SentiHoodTA} contains sentences extracted from a question answering platform where the main topic is urban neighbourhoods. As stated by the authors, Sentihood ``extends both aspect-based sentiment analysis that assumes a single entity per document and targeted sentiment analysis that assumes a single sentiment towards a target entity''. Like the Baby Care dataset, in Sentihood, each sentence could also have more than one (entity-aspect: polarity) pair labeled. All entities and aspects are predefined.

%===========================================
\subsection{Discussion}
\label{ssec:data-discussion}
%===========================================

%%
\begin{figure}
	\centering
    \subfigure[Review site\label{sfig:sitereview}]{\includegraphics[height=1in]{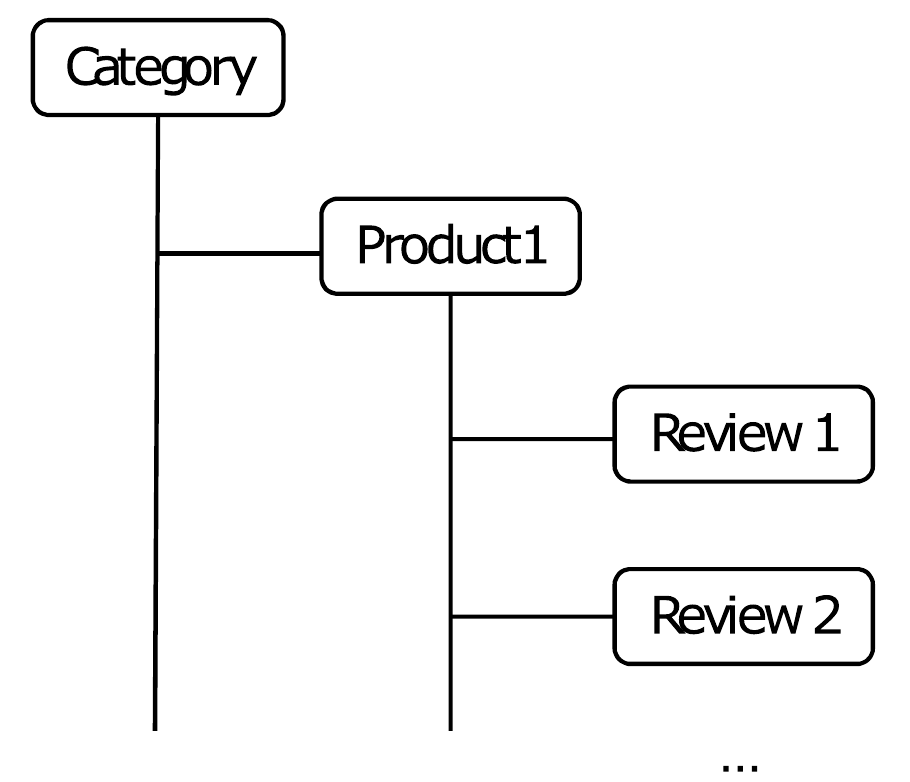}}
    \subfigure[Forum site\label{sfig:siteforum}] {\includegraphics[height=1in]{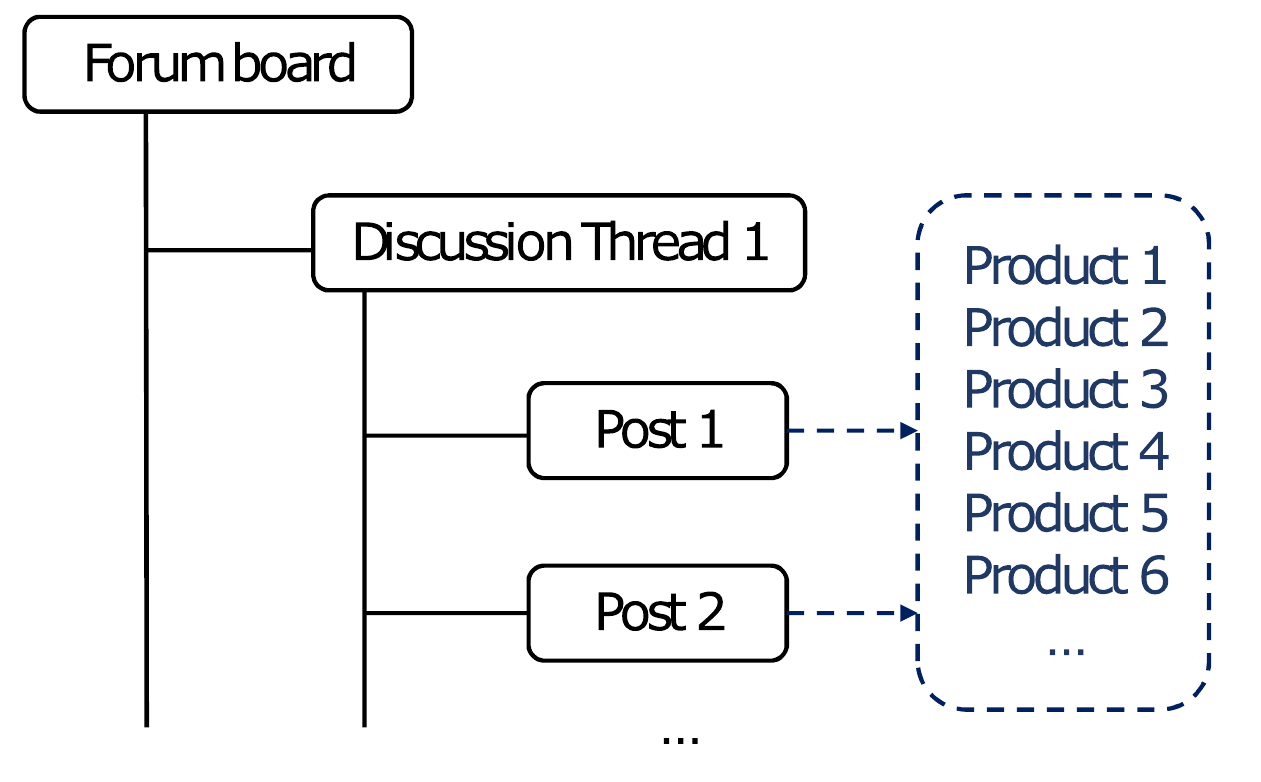}}
	\caption{Typical structure of review sites and forum sites	}
\label{fig:sitestructure}
\end{figure}

 SemEval 2014  is a typical product/service review dataset in which the target entity for all the sentences is implicitly given (\eg a restaurant or a laptop). In fact, many online services host a profile page for a product (or service) for users to comment on. Examples include products in Amazon, hotels in TripAdvisor, restaurants in Yelp, places in Foursquare, and many others. Although the target entity may not appear in each sentence and parts of the entity could also be considered as a target entity, all the reviews are considered targeted on a specific product/serice.\footnote{There could be cases where a review compares similar products or services. We consider that the review sentences on the target entity dominate a review text.} Further, the products are often grouped into predefined categories such that all products within the same category share similar attributes or properties, as shown in Figure~\ref{sfig:sitereview}. In this sense, it is relatively easy to identify the common aspects on which users may comment on, \eg location, service, food, and price for restaurants.

 Online forums, including question/answer services, typically are domain-specific. That is, the discussions within a forum (or a board of discussion to be more specific) are on similar domain-specific topics. As illustrated in Figure~\ref{sfig:siteforum}, all entities that are relevant to this topic are frequently mentioned, such as diaper and milk powder of different bands in a baby care forum. In this setting, the common characteristics of a group of similar products (\eg diaper) are relatively easy to be identified. However, in forum discussion, sentences in the same forum post may  target on different entities \eg diaper of different brands, or diaper and its related products. This is different from that in product reviews where all the sentences can be assumed to be commenting on the specific product to which the review was written.

 Twitter, similar to many social media platforms, provide a channel for common users to express their opinions \textit{on any topic}. Because of the extremely broad topic coverage, for sentiment analysis towards any specific entity (\eg person, organization, product, or event), relevant tweets have to be first collected (and filtered) from the Twitter stream. For example, the tweets in~\cite{dong2014adaptive} presented earlier were collected through keyword queries. Because of this task-specific collection process, the collected tweets likely contain sentiments expressed to the relevant entities. However, the sentiments expressed may not be easily grouped into aspect categories.

%===================================
\section{Categorization of TSA}
\label{sec:categorization}
%===================================

Targeted Sentiment Analysis (TSA) generally refers to a series of tasks aiming at detecting sentimental polarities towards targets in a given sentence. Here, the `target' could be entities, aspects, or entity-aspect pairs.  According to the different hierarchical structure of entity and its aspects for the sentiment target, we categorize TSA into three sub-tasks, namely, \textit{Target-grounded Aspect-based Sentiment Analysis} (\textbf{TG-ABSA}), \textit{Targeted Non-Aspect-based Sentiment Analysis} (\textbf{TN-ABSA}), and \textit{Targeted Aspect-based Sentiment Analysis} (\textbf{T-ABSA}). TG-ABSA applies to the product reviews where the target entity is explicitly or implicitly pre-defined. TN-ABSA applies to tweet-like documents where the targeted entities are pre-determined through  relevant document collection, and the opinions on these entities may not be easily grouped into aspects. T-ABSA is applicable to forum discussions where a group of similar entities are discussed or compared on a number of aspects.

%=====================================
\subsection{Target-grounded Aspect-based Sentiment Analysis (TG-ABSA)}
%=====================================

For TG-ABSA, the object of the analysis is the aspect of predefined targeted entity. A sentiment target $t$ could be described by ($e_0$, $a$) where $e_0$ refers to the predefined target entity for which the opinion document was composed, and $a$ is an aspect corresponding to $e_0$. Since all the aspects share the same target entity $e_0$, we could also describe the sentiment target $t$ as the target aspect $a$ for simplicity. A large number of aspect-based sentiment analysis tasks in the previous studies are TG-ABSA based on our  classification.

\paratitle{Aspect Extraction.} In the aspect-based sentient task proposed for SemEval 2014 task 4,  \citeauthor{Pontiki2014SemEval} highlighted two concepts: \textit{aspect term} and \textit{aspect category}. Aspect term refers to aspects explicitly named in a given sentence while aspect category is coarser and may not be named. However, such categorization is kind of problematic as a word like ``price" could be either aspect term or aspect category. Besides, in~\citeauthor{Liu2015SentimentAM}'s book, the concept ``aspect category" is used to refer to ``a unique aspect of the entity" as a same aspect could be expressed by different words. More importantly, whether a list of aspects is predefined or not is vital for real applications. Therefore we propose to categorize the aspect extraction task according to the existence of a predefined list of aspects.

To build a system for TG-ABSA, the first step is to identify a list of aspects to be targeted on for the predefined (types of) entities. This process could be completed either manually or by an algorithm.  Example aspects like \textit{food}, \textit{service}, \textit{ambience}, and \textit{price} could be identified for restaurant reviews. In many dedicated review websites, a predefined aspect list is provided for users to rate on. After identifying a list of aspects, a subtask in TG-ABSA is to extract the aspects (and may also include the aspect terms) indicated in review sentences.

Both supervised and unsupervised methods have been proposed for aspect extraction. Supervised methods usually formulate aspect extraction as a sequence labeling task.  \citeauthor{Katiyar2016InvestigatingLF} investigated the use of deep bi-directional LSTMs for joint extraction of opinion entities and the IS-FORM and IS-ABOUT relations that connect them. \citeauthor{Wang2017CoupledMA} offered a multi-layer attention network for co-extraction of aspect and opinion terms where each layer consists of a couple of attentions with tensor operators. \citeauthor{Ding2017RecurrentNN} proposed to combine rule based methods with neural models for cross-domain opinion target extraction. Unsupervised methods consider aspect extraction as a representation learning problem and usually focus on the learning  of word embeddings. \citeauthor{Yin2016UnsupervisedWA} extracted aspect terms by learning distributed representations of words and dependency paths. \citeauthor{He2017AnUN} proposed an attention-based model for aspect extraction by focusing more on aspect-related words while de-emphasizing aspect-irrelevant words during the learning of aspect embeddings.

\paratitle{Polarity Identification.}  In TG-ABSA, polarity identification is to determine the polarity of aspect terms in a sentence with respect to the given  aspects. For instance, in a sentence \textit{``I liked the aluminum body.''}, the aspect term is ``aluminum body'' and the polarity is positive.  Again, in TG-ABSA, the target of the review is predefined, which is a laptop for the above example sentence.

Many solutions for polarity identification proposed in recent studies are based on memory network.  \citeauthor{tang-qin-liu:2016:EMNLP2016} firstly introduced deep memory networks into aspect level sentiment classification task. \citeauthor{chen2017recurrent} adopted multiple-attention mechanism to capture sentiment features separated by a long distance based on memory networks. \citeauthor{tay2017dyadic} proposed a novel extension of memory networks by enabling rich dyadic interactions between aspect and context. \citeauthor{wangACLtarget} improved the memory networks by adding interactions between the context and the target.  \citeauthor{zhu2018enhanced} devised a novel deep memory network with auxiliary memory and the features of aspects and terms could be simultaneously learned via interactions between the two memory networks.

There are also solutions based on recurrent neural network.  \citeauthor{AAAI18wordaspect} incorporated aspect information into the neural model by modeling word-aspect relationships. \citeauthor{wang2018aspect} proposed a hierarchical network with both word-level and clause-level attentions. \citeauthor{Wang2018LearningLO} implemented CRF (conditional random fields) based on attentive LSTM which is able to  extract interpretable sentiment expressions. \citeauthor{gu2018position} proposed a position-aware bidirectional attention network which not only concentrates on the position information of aspect terms, but also mutually models the relation between aspect term and sentence by employing bidirectional attention mechanism. \citeauthor{he2018effective} proposed two  approaches which could improve the effectiveness of attention for target sentiment analysis.

Other than memory networks and RNN, \cite{Nguyen2015PhraseRNNPR} proposed a Phrase Recursive Neural Network that takes both dependency and constituent trees of a sentence into account. \cite{Xue2018AspectBS} leveraged convolutional neural networks and gating mechanisms to build more accurate and efficient model. \cite{Barnes2016ExploringDR} explored distributional representations and machine translation for aspect-based cross-lingual sentiment classification~\cite{Akhtar2018SolvingDS} proposed to leverage bilingual word embeddings to solve the data sparsity problem and have achieved state-of-the-art performance in two experimental setups. \cite{He2018ExploitingDK} explored two approaches to transfer knowledge from document-level data to TG-ABSA task.

%=====================================
\subsection{Targeted Non-aspect-based Sentiment Analysis (TN-ABSA)}
%=====================================

TN-ABSA is usually named as Target-dependent Sentiment Analysis or Entity-level Sentiment Analysis in previous studies. The data is usually collected from social media like twitter, and it is usually hard to define common \textit{aspects} in such documents. For TN-ABSA, the object of the analysis is simply the target entity. Thus the sentiment target $t$ could be described as the target entity $e$.

A key subtask in TN-ABSA is entity extraction, which is to identify the target entity mentioned in the given sentences. This subtask is basically named entity recognition (NER) which has been heavily studied in NLP and IR communities~\cite{Yadav2018ASO}.

For the subtask of polarity identification, existing approaches for TG-ABSA and TN-ABSA are quite similar. Works like \cite{li2018transformation} are tested on both Twitter dataset and restaurant dataset in SemEval 2014 Task 4. \citeauthor{dong2014adaptive} proposed adaptive recursive neural network which could model structural dependency in context. \citeauthor{tang-EtAl:2016:COLING3} devised two target dependent LSTM models to capture interactions between target and context. \citeauthor{zhang2016gated} built gated neural networks to generate better representations for target and surrounding text.  \citeauthor{yang2017attention} presented an attention-based bidirectional LSTM approach to improve the target-dependent sentiment classification.  \citeauthor{li2018transformation} combined CNN and RNN to build a transformation network and achieved state-of-the-art performance on two datasets.

    %and  The very reason why we divided them into two tasks is their applications and data collection process are totally different.

%=====================================
\subsection{Targeted Aspect-based Sentiment Analysis (T-ABSA)}
%=====================================
For T-ABSA, the object of the analysis is the entity-aspect pair. Thus the sentiment target $t$ could be described as ($e$, $a$). Both entities and aspects may explicitly present in the opinion document. T-ABSA usually happens in online forums with a specific topic like camera, mobile phone, or fun places around a city. If the topic is too specific like a single thread created in a discussion board solely for one product, then the task becomes similar to TG-ABSA.

Two subtasks, target extraction and polarity identification can be formulated for T-ABSA. However, existing works only focus on analyzing sentiment expressed on pre-annotated (entity, aspect) pairs.\footnote{The correct identification of (entity, aspect) pairs is considered more challenging than entity identification and aspect identification.} Here we only briefly review solutions proposed for polarity identification in T-ABSA.  \cite{Saeidi2016SentiHoodTA} could be considered as the first attempt on this task with the introduction of the Sentihood dataset. \citeauthor{Ma2018TargetedAS} proposed to integrate commonsense knowledge into attentive LSTM. \citeauthor{Liu2018RecurrentEN} proposed recurrent entity networks with delayed memory update and achieved state-of-the-art performance on Sentihood dataset. \citeauthor{yang2018multi} built memory networks for context, entities, and aspects respectively and further devise interaction and attention layers to extract target dependent representation for the given opinion documents.

%=====================================
\subsection{Discussion}
\label{ssec:models}
%=====================================

 \begin{figure}
    \centering
    \includegraphics[width=0.9\columnwidth]{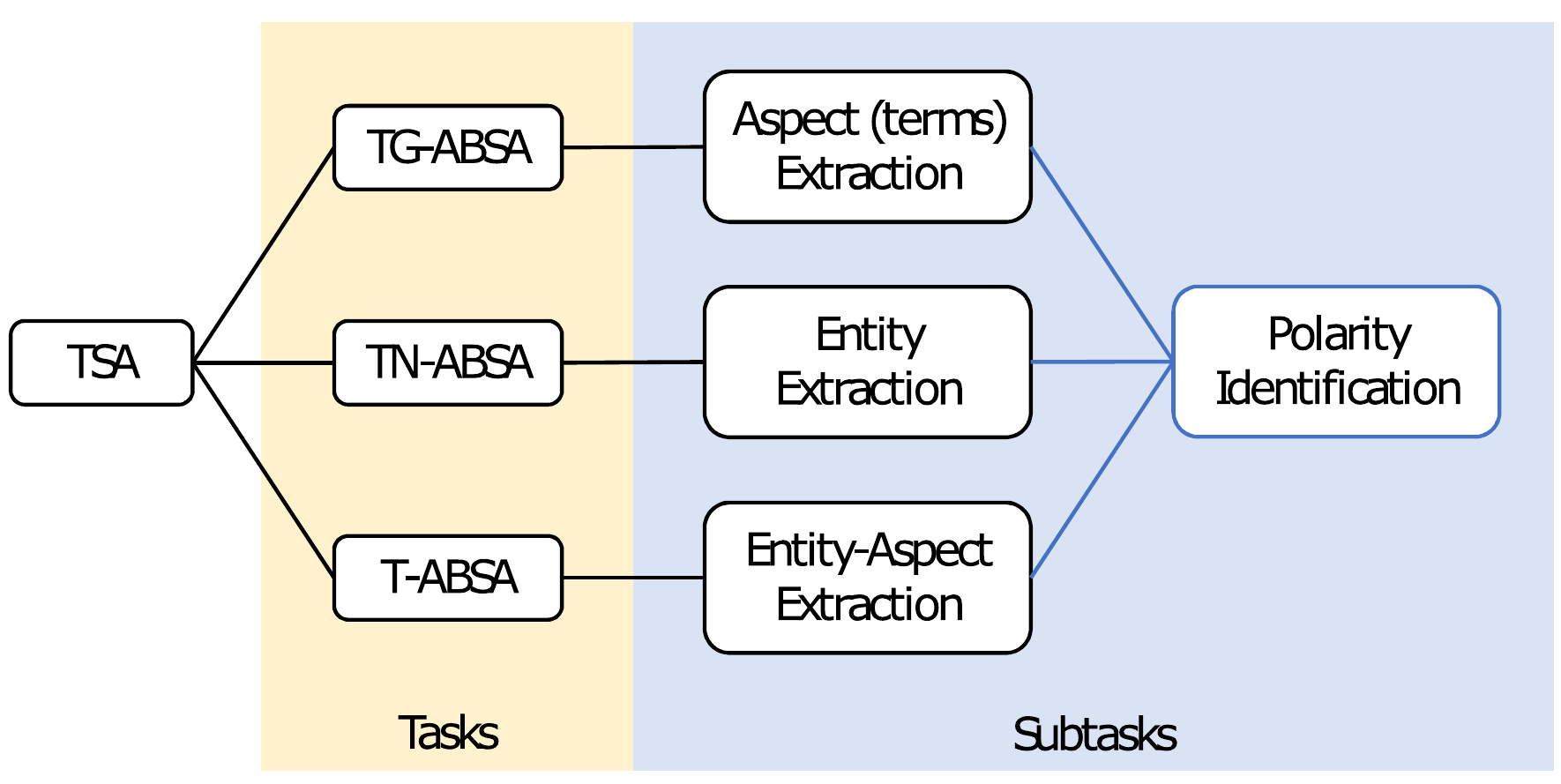}
    \caption{Categorization of TSA tasks}
    \label{fig:hierarachy}
 \end{figure}

\begin{table}
\centering
{\small
\caption{Summary of recent works for TSA}
\label{tab:tsawork}
\begin{tabular}{@{}l|ll@{}}
\toprule
Paper & Sentence Rep. & Target-Context  \\
\midrule
\cite{tang-qin-liu:2016:EMNLP2016}    & Memory Network          & Attention\\
\cite{chen-EtAl:2017:EMNLP20171}     & Memory Network          & Attention \\
\cite{tay2017dyadic}      & Memory Network          & Dyadic Interaction      \\
\cite{wangACLtarget}      & Memory Network          & Built-in          \\
\cite{wang2018aspect}      & LSTM        & Attention         \\
%\cite{Wang2018LearningLO}  & LSTM        & CRF               &  \\
\cite{Xue2018AspectBS}     & CNN         & Gate operation \\
\cite{gu2018position}  & GRU             & Attention                \\
\cite{he2018effective}  & LSTM           & Attention                \\
\cite{zhu2018enhanced}  & Memory Network & Attention  \\
\cite{tang-EtAl:2016:COLING3} & LSTM     & Built-in           \\
\cite{zhang2016gated}     & Gated RNN    & Built-in           \\
\cite{yang2017attention}  & LSTM         & Attention          \\
\cite{li2018transformation} & LSTM       & CNN               \\
\cite{Ma2018TargetedAS} & LSTM           & Attention        \\
\cite{Liu2018RecurrentEN} & GRU          & Gate operation    \\
\cite{yang2017attention}  & LSTM         & Attention          \\
\bottomrule
\end{tabular}
}
\end{table}

We summarize the three types of TSA tasks, \ie TG-ABSA, TN-ABSA, T-ABSA and their subtasks in Figure~\ref{fig:hierarachy}. Through the review of existing solutions, we note that the studies on the subtask of target extraction in these three tasks are relatively limited. In fact, in many studies, the targets (\eg aspects in TG-ABSA and entity-aspect pairs in T-ABSA) are assumed to be available and pre-annotated in the given documents. Because the `availability' of targets, a large number of studies for TSA focus on polarity identification of the given targets. Recent deep learning models designed for TSA  usually focus on \textit{how to represent a given sentence} and \textit{how to build better interaction between the target and its context}, summarized in Table~\ref{tab:tsawork}.
For different subtasks of TSA, specific solutions to solve the two problems may be slightly different. However, representation and interaction are always key factors to the success of TSA tasks.

\paratitle{Sentence Representation.} Words are firstly embedded into dense representations which are then fed into the sentence representation model. Recurrent neural networks (LSTM or GRU) and Memory Networks are widely adopted in existing TSA architectures to model the entire sentence~\cite{yang2017attention,Ma2018TargetedAS}. A straightforward way to use RNN is feeding each word into the RNN model at each time step. Then the hidden states at the last time step could be considered as the representation for the entire sentence. RNN could model long term dependency and retain structural information in the given sentence. However, RNN may lose original information captured by individual words.
%\textbf{Memory Networks: }
For memory networks, a simple way is to put word embeddings into memory slots. Memory networks could retain all original word-level information. However, it loses structural information and long term dependency. Thus position embedding~\cite{gu2018position} and multi-layer attention mechanism~\cite{tang-qin-liu:2016:EMNLP2016,chen-EtAl:2017:EMNLP20171} are usually used with Memory networks.

\paratitle{Target-Context Interaction.} Apart from sentence representation, TSA models also need to model interactions between the target and its context. In some works, context interactions are built within the sentence representation model. For example in TD-LSTM~\cite{tang-EtAl:2016:COLING3}, two LSTMs are devised to model right and left word sequences with the target respectively. \citeauthor{wang2018aspect} built target sensitive memory networks and each item in memory networks is saved after interactions between the target and individual words.

In many other works, attention is  adopted to generate target dependent representation~\cite{yang2017attention,zhu2018enhanced}. A straight way is to use target embedding as the attention query thus the highest weights could be assigned to the most relevant context words. CNN~\cite{li2018transformation}, gate operation~\cite{Xue2018AspectBS,Liu2018RecurrentEN}, and other techniques could also be devised to model the interactions between target and context.

%=====================================
\section{The Data to TSA Tasks}
%=====================================

\begin{table*}
\centering
\caption{Categorization of the data}
\label{tab:autometrics}
{\small
\begin{tabular}{@{}l|p{0.72in}|c|c|c|c|l@{}}
\toprule
Task & Dataset & Coherence  & Source & Collection & Target Structure & Example application domain\\
\hline
TG-ABSA & SemEval 2014 & Strong & Online Review & Crawling & Aspect (Entity) & Product, service, movie, Apps   \\
TN-ABSA & Twitter & Weak & Twitter & Filtering & Entity & Event, people, organization   \\
T-ABSA & Sentihood, Baby Care & Moderate & Forum& Crawling & (Entity, Aspect) & product, service   \\
\bottomrule
\end{tabular}
}
\end{table*}

Data is always one of the key issues to all machine learning applications including sentiment analysis. We categorize existing datasets to TSA tasks from the perspective of their intrinsic characteristics and the collecting process. The categorization of the datasets is consistent with the categorization of TSA tasks (\ie TG-ABSA, TN-ABSA, T-ABSA) as the tasks are defined based on the information available in the datasets. However, the data collection and annotation process are often ignored in the papers reporting advances made in TSA, making the proposed models less applicable in real scenarios. The dedicated collection process and expensive annotation process may also introduce a gap between the datasets for evaluation purpose of TSA tasks and the data to be dealt with in real world applications.

%=====================================
\subsection{Data Categorization}
%=====================================
We characterize the datasets applicable to the three TSA tasks from the following three perspectives.

\paratitle{Topic Coherence}  describes the inner similarity of documents in different datasets and it is the intrinsic difference between different datasets and tasks. For datasets collected from product/service review sites, opinion pieces have the strongest topic coherence and relevance as they are all on a same target entity. Because of such strong topic coherence, it is possible that fine-grained aspect information could be derived or determined. For data collected from social media platforms, the data demonstrates less topic coherence, \ie opinions are not on common aspects of the entities of interest.  For online forums, the discussions are in general relevant to a specific topic but the discussion could be much diverse compared with dedicated review sites. Further, as the discussions are not specific on a predefined entity but on a collection of similar entities, the focus (or the aspect) of the discussion evolves along time depends on the nature of product. For example, new features are introduced to smartphones every couple of months.

\paratitle{Target Structure} generally defines each specific task in TSA. If the goal is to analyze different aspects of a single product, then the data has to be prepared with such information available in the data.  If the goal is to understand general sentiment polarity toward some entities, the data could be collected from more open domain platforms. The target structure determines the amount of pre-processing work before applying the existing solutions that have  been proposed for TG-ABSA, TN-ABSA, and T-ABSA respectively. We note that in many existing studies, the targets are pre-annotated.

\paratitle{Source \& Collection: } As the most studied task, data for TG-ABSA are typically collected from online shopping platform or review websites. For data collection from social media like Twitter, information filtering is a major concern to select only the relevant tweets. Note that, a tweet containing a relevant keyword does not imply that the tweet is relevant to the entity of interest due to many factors, such as language disambiguation, spam, spelling variations.

%=====================================
\subsection{Challenges}
%=====================================

The categorization of data (and also the three TSA tasks)  provides a reference path on how to build a practical application on real data. Developers may determine the required data resources and the subtasks to be addressed in applying specific TSA models. However, the available benchmark datasets are far from enough to be used in many real applications as these datasets were built  by fitting  predefined tasks. Many TSA models proposed are only applicable on data with pre-annotations, \eg the entity-aspect pairs.  Therefore there is a big gap in applying the models on real data where the pre-annotations are unavailable. For example, it is infeasible to manually annotate the aspects for a large number of products even the data are collected from review websites.  Another example is public opinion tracing where collecting and filtering event-relevant data from social media itself is not well addressed yet. Overall, we consider two major challenges related to the data to TSA tasks.

Where to collect the appropriate data is the first challenge. Twitter, forums, news streams, and custom feedbacks  are all potential data sources in addition to dedicated review websites. However, intrinsic features of the data and the filtering process have significant impact on the applicability of the models and their performances.  Specifically, the filtering process in identifying the relevant data become the key challenge. There are a lot of practical issues to be solved based on the real application requirement. For example, if a company aims to analyze public opinions towards a recent event from tweets, then collecting the event-relevant tweets from Twitter stream becomes the key requirement. Filtering through keywords, hashtags, and even geographical indicators has limited filtering power. Further, duplication and near-duplication identification is another issue to be addressed. The same applies to data collection from news stream. Although data collection is not considered the subareas in sentiment analysis, the topic coherence and relevance affects the applicability of models and their performances.

After data collection, in the data annotation process, one needs to determine not only the ``target'' (\eg entity, aspect, or entity-aspect pairs) but also the right ``context'' which containing sentiment on these targets. As we discussed in Section~\ref{ssec:models}, modeling the interaction between target and context is a key issue in TSA, and existing studies all assume that the context are provided. In reality, the context may not be well defined. For example, if tweets are the documents of interest, then each tweet can be considered as the context containing sentiment on a target mentioned in the tweet. However, for a news article, an entity (or target in this context) may be mentioned in a few sentences in the entire article, then the issue is how to extract the relevant sentences expressing sentiments towards this target. In other words, only a small part of the entire news article is relevant and providing the context of sentiment analysis.

%=====================================
\section{Conclusion}
%=====================================

In this paper, we categorize target sentiment analysis tasks and the datasets reported in recent years, and summarize the solutions. We discuss the differences between the three categories: TG-ABSA, TN-ABSA and T-ABSA, to provide a fine-grained understanding of TSA, including their subtasks. This categorization clarify the different definitions of targets, \ie entity, aspect, and entity-aspect pair. We then discuss the differences as well as the requirements on the data to be fed into the different TSA models. Our categorization could be used as a guideline to build a TSA system, from data collection to model design. Moreover, we also discuss challenges for TSA tasks, particulary from the data requirement perspective, which are not detailed in many recent studies.